# An Artificial Neural Network Architecture Based on Context Transformations in Cortical Minicolumns

**Vasily Morzhakov, Alexey Redozubov**
morzhakovva@gmail.com, galdrd@gmail.com

## Abstract

Cortical minicolumns are considered a model of cortical organization. Their function is still a source of research and not reflected properly in modern architecture of nets in algorithms of Artificial Intelligence. We assume its function and describe it in this article. Furthermore, we show how this proposal allows to construct a new architecture, that is not based on convolutional neural networks, test it on MNIST data and receive close to Convolutional Neural Network accuracy. We also show that the proposed architecture possesses an ability to train on a small quantity of samples. To achieve these results, we enable the minicolumns to remember context transformations.

## 1. Introduction

Human perception always deals with ambiguity in the brain. Only receptors provide us with clear input. As the input data from receptors start to be analyzed, explicit treatment disappears. Almost every frame of data can have a different sense in a different context. This is reflected in all types of perception. There are many illusions in visual perception that are caused by the presence of different geometric contexts, for example: "inverted face". Our language seems to have to exclude the ambiguity, but it doesn't. All humor is based on unexpected changes of contexts. Even our delusions are possible because of our ability to find context where they have sense. All these specificities of human perception are not fundamentally reflected in modern machine-learning approaches.

Artificial neural networks (ANN) are presented by different architectures today. Recurrent ANNs and Convolution ANNs don't have much in common and are based on different concepts. Meanwhile, the neocortex is very homogeneous through its surface. This means that a new concept has to be applied for different types of data: images, video, audio and natural language. Even manipulators' controlling should be explained in this approach. Now, all these tasks use completely different algorithms and theories.

One more fundamental problem of modern ANNs is that they need more samples for training than humans. Convolution ANNs (CNN) take into account translation invariants, recurrent ANNs (RNN) – time invariants. It's impossible to know how rotate objects will be described by a CNN without showing a particular object in different rotations. RNN can't extract a twice slower process in time if it isn't presented in training data. CNN and RNN were breakthroughs because they could extract features in different positions and time. Position and time are contexts too. We also deal with a lot of other contexts all the time: orientation, size, frequency, tempo, shape and others. We have awareness of how the data would be changed in all of those contexts. It will dramatically reduce the amount of samples we need for extracting features in the input data we have.

These facts allow us to find new concepts that will allow us to go beyond all limitations described above.

In this article, we propose function of minicolumns in the neocortex and describe a sample implementation for MNIST data that is based on actual machine-learning framework.

### 1.1 Brain zone and minicolumns functions

Hubel and Wiesel [2] proposed a concept that is widely accepted in modern implementations of Convolutional Neural Networks (CNN). They claimed that brain zones in the neocortex are organized into minicolumns that contain neuron-detectors with similar functions. They found that in the visual zone V1, minicolumns have orientational selectivity in their experiments. A lot of research shows different types of selectivity in different brain zones. According to these experiments, a few approaches in computer vision appeared: self-organizing maps and neocognitron. The neocognitron architecture found its realization in convolutional neural networks that still provide the best accuracy of recognition in tasks of computer vision. Self-organizing maps are also well-known in tasks of unsupervised learning.

We propose another interpretation that allows the creation of an original approach for data generalization and recognition tasks, inspired by more relevant information to today's research. Roots of our interpretation are described in detail in [1].

Almost all data processed by the brain is presented unambiguously. Objects are always projected to the retina in different ways, words are pronounced in different tones, even words in our language have different senses in different contexts. We assumed that each minicolumn in the neocortex presents the input information in its own context. The second task of the minicolumn should solve pattern recognition. Activity of minicolumns is proportional to certainty of recognition in the context of each minicolumn. Close minicolumns represent similar context transformation. This way we can single out local maximums of activity in the field of minicolumns and send the following output to the next level: what the minicolumn has recognized and in what context (transformation). Also, it's important that patterns for recognition are shared among minicolumns.

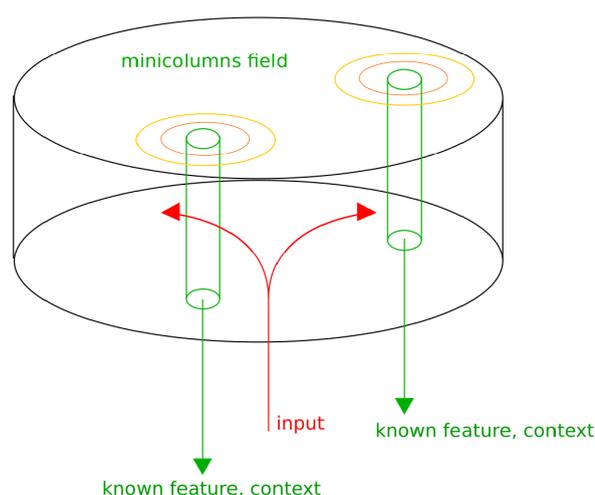

Figure 1, Input data received by a zone is translated in different contexts by minicolumns, a few of minicolumns send what they recognize and its contexts to the following level

Thus, neurons should have two types of memory: memory for transformation and memory for generalization. Modern vision of neurons' structure has gone far from the vision in times of Hubel and Wiesel. Possible mechanisms of these types of memory will be described in [1]. The sample architecture in this article is based on the TensorFlow framework and the classic perceptron conception.

## 2. Model for MNIST

MNIST dataset was chosen in order to demonstrate described functionality of minicolumns inside one zone.

We created a model [7], that doesn't contain all features of our approach, but which should illustrate proposed functions of a brain zone. This sample model was based on well-developed frameworks for deep learning, thus the architecture and source codes are transparent for deep learning specialists.

A zone receives a compressed vector (code) that describes relevant information for the zone. We trained an autoencoder with a bottleneck to obtain the input vector. Additionally, we were able to use its decoder part for visualization of codes in minicolumns. The green layers in figure 2 mark encoder and decoder parts of the autoencoder.

As in 1.1, described transformation to different contexts should be remembered. We chose geometric transformations for MNIST images: translation, scale, rotation. 405 minicolumns were set ( 9 translations x 9 scales x 5 rotations = 405 transformations).

Transformation remembering was implemented by ANNs with one hidden layer for each. Each ANN was trained to transform the input vector to the code of the transformed image.

After input codes were transferred into different contexts, we had codes in each minicolumn. Thus, 405 minicolumns contained codes of images if they were geometrically transformed. These codes are decoded by the decoder part of the autoencoder for visualization (2.2).

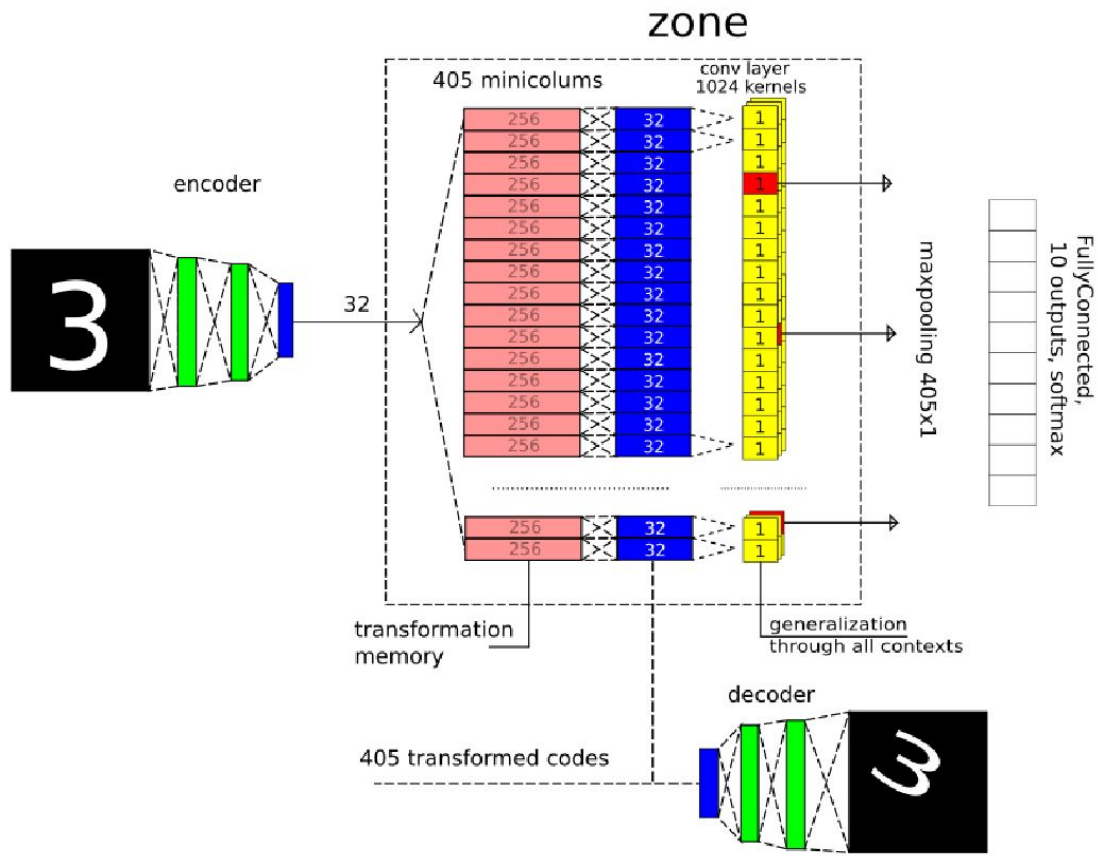

Figure 2, model for MNIST data,
green layers – autoencoder part
red layers – transformation part
yellow – generalization with convolution layers (weights are shared for transformed codes)
blue – codes, that describe images

The next step was feature extracting from the minicolumns. The best way for this was using the convolution layer. Convolution layers have kernels (or features), the weights for detecting those features are applied for each position on an image. So, minicolumn codes might be presented as an image that allow the application of the same weight vector for each minicolumn. In this way, we could extract features despite their context. It allows for a decrease in the number of weight coefficients in CNNs and an increase in the quality of prediction. Following the function of a brain zone, the output equals the local maximums of the last convolution layer. For this sample, we left only one maximum for each channel and applied a max-pooling layer. The output of the max-pooling layer was connected to a fully-connected layer with 10 outputs and softmax activation for training recognition on MNIST.

## 2.1 Autoencoder

First of all, we needed to extract a vector that would be an input for the zone. An autoencoder with two fully-connected layers allowed to compress input (28,28) images to vectors in 32 dimensions.

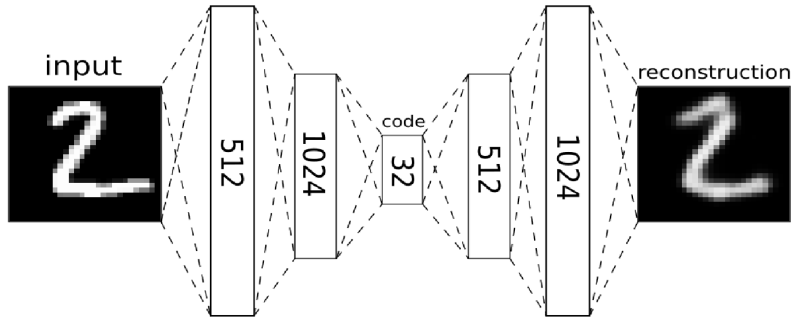

Figure 3, An autoencoder with fully-connected layers

Two fully-connected layers were used for encoding an input image and two for decoding it. The network was trained with geometrically augmented images from MNIST.

Mean squared error was chosen while training. The mean squared average error was 0.011 after 50 epochs. These results were a bit worse than the error of a convolutional autoencoder in our experiments (0.006), but it was very important to exclude the CNN for the purity of experiments.

## 2.2 Context transformation training

The next stage was training context transformations. In the article we use the term "context transformation training" because we used the back-propagation approach, however, this process should be closer to remembering.

In some cases it's possible to define a set of transformation. For example, for MNIST dataset, we could define geometric transformations: translation, scale, rotation. There were 405 minicolumns for 405 geometric transformations. We trained a separate model (figure 4) for each context resulting in 405 nets trained one by one. While training, we entered two codes into the model: before and after transformation.

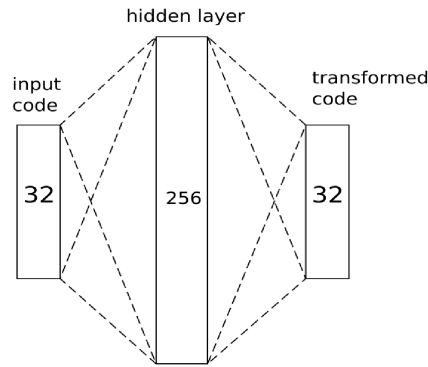

Figure 4, context transformation model for a minicolumn

Input images were encoded into short codes. Each code was transformed to 405 vectors by trained nets, that could be decoded by the second part of the autoencoder (decoder part) and visualized:

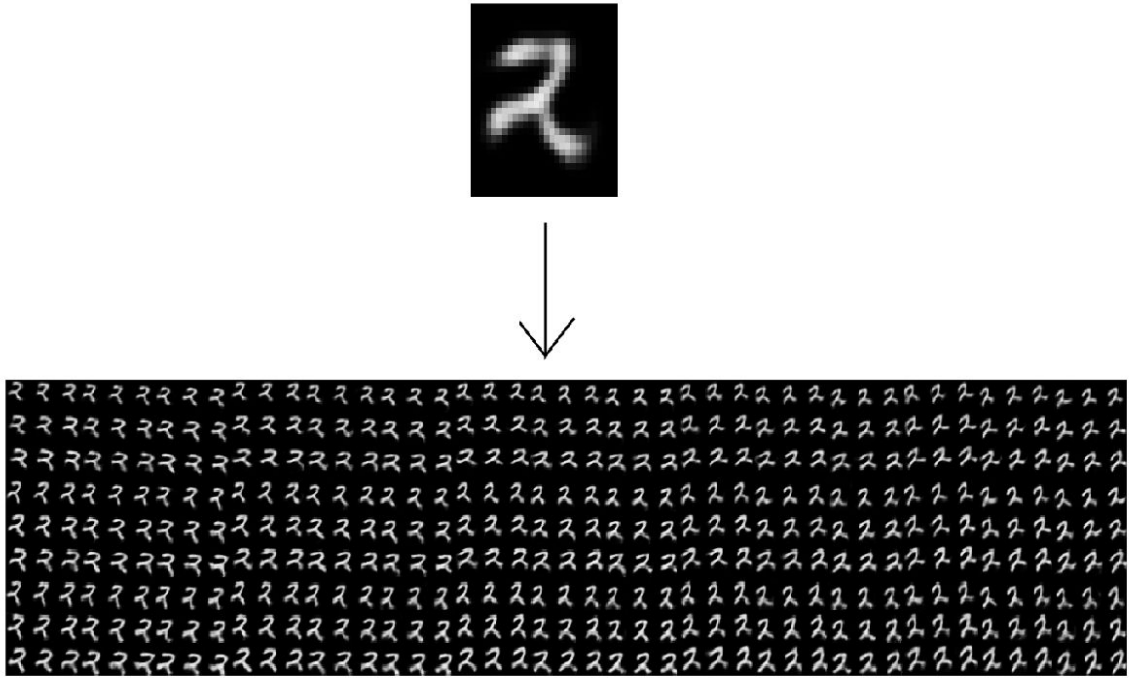

Figure 5, a sample of decoded codes after transformations in 405 minicolumns visualized by the decoder

Figure 5 shows that all of minicolumns were trained successfully. After decoding codes in minicolumns, we could see transformed input images.

Additionally, this approach of training transformation has a remarkable prediction ability. For example, in figure 6, transformations of a triangle standing on its edge are shown:

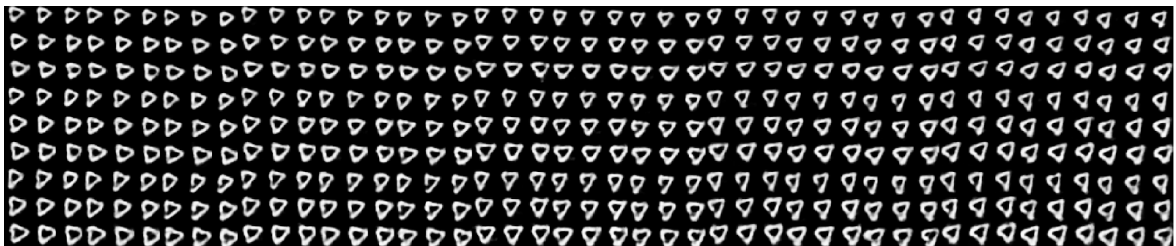

figure 6, transformations of a triangle sign

This sign resembling digits was not presented in the training set. This meant that nets could predict transformations for code they have never seen. It offers hope of a finite training set for remembering transformations of full a variety of input codes into different contexts.

### 2.3 Recognition whether contexts

After we received the codes from the minicolums, we could apply the same detector for each minicolumn for pattern recognition.

Different approaches may be applied for detecting patterns in different contexts, it could be remembering samples. The only thing that should be defined is context presentation choice for remembering. For this reason, we decided to use the back-propagation mechanism for training detectors' weights.

The shortest way for sharing detectors through different contexts was to use a convolution layer. Instead of sharing weights among detectors at different positions, the convolution layer provided weight sharing for different contexts. This feature is especially important because it allows recognition patterns even if there were no samples in all representations, therefore, "one shot learning" becomes possible. Then, for each channel a max-pooling layer is applied that actually chooses the best matching context.

For MNIST data recognition, we needed one additional layer – a fully-connected layer with softmax activation with 10 output neurons.

While the training procedure weights in the autoencoder and the transformation net was frozen, only the weights in the convolutional and output fully-connected layers were trained.

## 3. Experiments

We tested the model on MNIST data. There were a few experiments:
1) Training on the first 1,000 images
2) Training on the whole dataset of 60,000 images

We also compared our results with a convolutional neural network (CNN) in case of augmented data and original data.

### 3.1 Training on 1,000 images

It's very important to show that it demands less data for training for the proposed architecture, that's why its performance was estimated by training on the first 1,000 images from dataset.

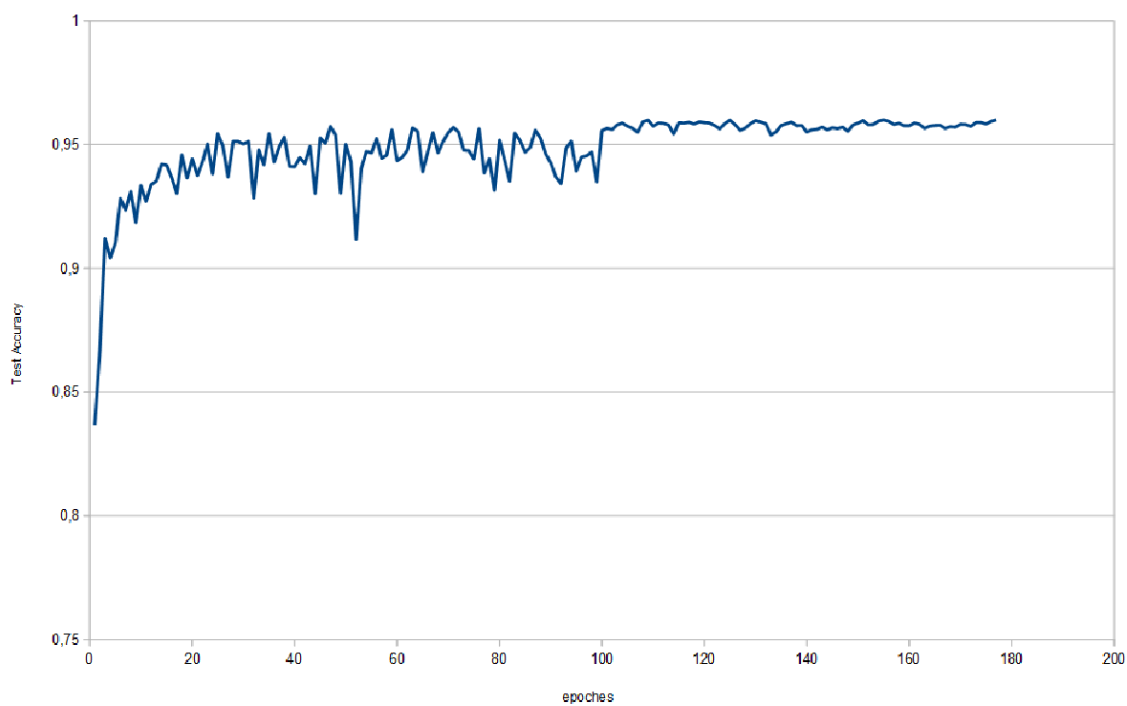

Figure 7, training curve for 1,000 images

After 180 epoches, we had 96.0% accuracy on 10,000 test images.

The baseline CNN for comparison had the following layers:

| Layer | Parameters |
| --- | --- |
| Convolutional Layer | 64 channels, 5x5 kernel size, ReLu activation |
| Max-pooling | 2x2 kernel |

| Convolutional Layer | 128 channels, 5x5 kernel size, ReLu activation |
| --- | --- |
| Max-pooling | 2x2 kernel |
| Fully-connected layer | 256 outputs, ReLu activation |
| Fully-connected layer | 10 outputs, softmax activation |

After training the baseline CNN, the accuracy was 95.3%

Once can see that we used augmented data implicitly while training context transformations. For this reason, the baseline CNN should be trained on augmented data, too. This test was also run resulting in 96.2% accuracy.

So, we can conclude that the proposed architecture shows comparable performance with the convolutional neural network, even without internal convolutional layers.

### 3.2    Training on 60,000 images

The best result was achieved after 1,000 epochs training on 60,000 images – 99.1%

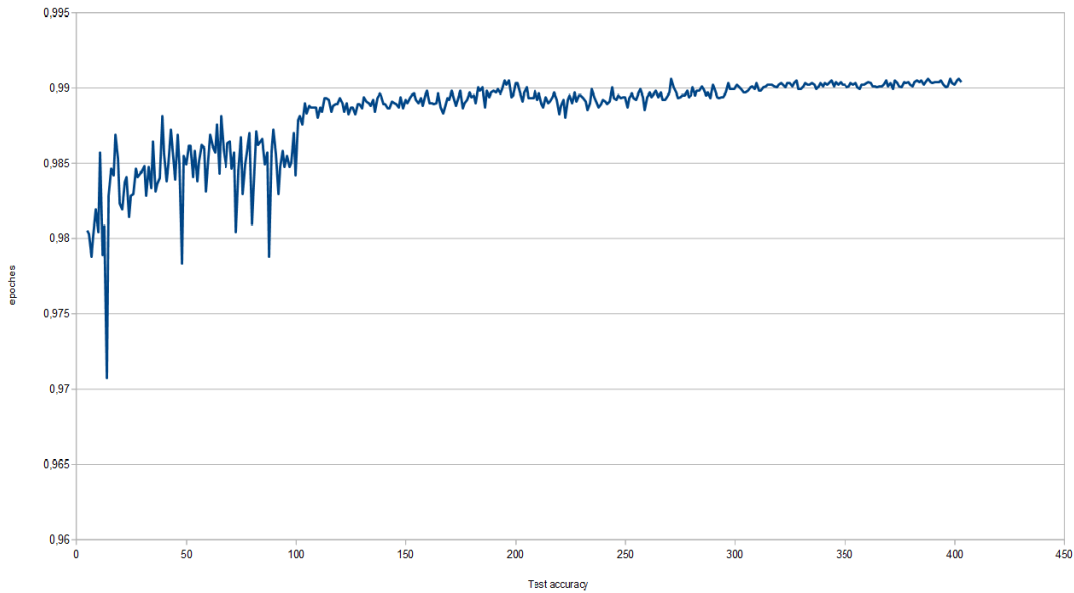

Figure 8, training curve for 60,000 images

This result is worse than what the CNNs achieve. It's supposed to be caused by presence of a not precise autoencoder that calculates short representation of data for input.

## Comparison to other approaches

The idea of image transformation in computer vision tasks is well known and described at least in Spatial Transformer Networks [5]. The approach authors proposal can be used across different types of transformations and types of data. The main difference is that it demands for differentiable transformation functions that can be unattainable in case of discrete data. Additionally, transformation cannot be trained in Spatial Transformer Networks.

Hinton writes about capsules that are inspired by cortical minicolumns in [3]. The proposed approach connects to the CapsuleNet in the "visual stream for perception". Objects can be contexts for recognition of transformations, but in the CapsuleNet, minicolumns don't remember how the geometric transformation changes in different contexts (digits in the case of MNIST data).

## Two-stream hypothesis

The Two-stream hypothesis is a well-known and accepted model of visual and auditory perception [4,6]. The ventral is called "what pathway", the dorsal stream "where pathway". Minicolumns in the first one are selective to object types, their shapes and colors. The second stream has minicolumns with position and orientation selectivity. [6]

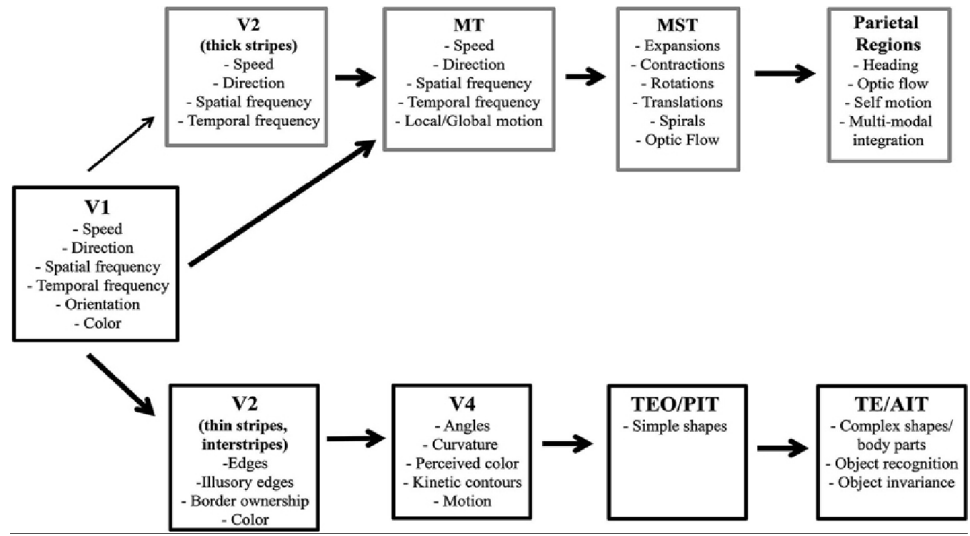

Figure 9, Hierarchy of visual processing in ventral and dorsal streams [6]

The interpretations that we propose change the way of perception in the visual stream of the brain significantly. Minicolumn activity results in the best context or transformation of the input data. Meanwhile, minicolumns in the "where pathway" contain sufficient information about objects inside to help recognize them in the context of a particular minicolumn.

It is quite intuitive in the case of the "where pathway". Contexts are just geometric transformations, as implemented in the MNIST sample implementation. It is not hard to imagine transformations mentioned in the upper part of Figure 7, because almost all of them are just changing the coordinate system in time and space. However, it's a bit counterintuitive when we speak "in context of an object". For example, in [3] Hinton has shown a model where the minicolumns' activation was connected with recognition of a particular digit in MNIST. The problem is regulated through the assumption that codes in minicolumns may characterize objects' properties. Those properties may be the same for different objects, for example, the font style for digits in MNIST. Thus codes in minicolumns during context transformation training will designate style. It is very close to the ideas of CapsuleNet, but styles' code will be shared among minicolumns.

The two-stream hypothesis provides the idea, that the context-transformation described in 2.2 may not translate input codes in the same space where input vectors were defined, but in a new space that describes something else.

## Conclusions

In this paper, we introduced a new architecture of artificial neural networks that is based on our original ideas of the cortical minicolumns' functionality. Transformation input data into a context specific to a minicolumn allows one to choose the best matching pattern while recognition. It also gives an opportunity for separation patterns and contexts in which they are presented. Moreover, there is no restriction for input data or for the type of transformation. As we discussed in the chapter "two-stream hypothesis" detected patterns in one zone may be contexts for other zones, that may give clues for "abstract thinking" of artificial intelligence.

## Acknowledgments

The authors would like to thank Anton Maltsev for his valuable remarks and discussions.